\def\year{2017}\relax
\documentclass[letterpaper]{article}

\usepackage{aaai17}
\usepackage{times}
\usepackage{helvet}
\usepackage{courier}

\usepackage[linesnumbered,ruled,vlined]{algorithm2e}

\SetCommentSty{mycommfont}
\usepackage{multirow}
\usepackage{graphicx}
\usepackage{amssymb}
\usepackage{subfigure}
\newtheorem{definition}{Definition}
\newtheorem{problem}{Problem}
\gdef\eg{\textit{e.g.}}
\gdef\ie{\textit{i.e.}}

\begin{document}
%
\title{Deep Spatio-Temporal Residual Networks for Citywide Crowd Flows Prediction\thanks{This research was supported by NSFC (Nos. 61672399, U1401258), and the 973 Program (No. 2015CB352400). }}
\author{Junbo Zhang$^1$, Yu Zheng$^{1,2,3,4}$\thanks{Correspondence author. This work was done when the third author was an intern at Microsoft Research. }, Dekang Qi$^{2, 1}$ \\ 
{$^1$Microsoft Research, Beijing, China}\\
{$^2$School of Information Science and Technology, Southwest Jiaotong University, Chengdu, China}\\
{$^3$School of Computer Science and Technology, Xidian University, China}\\
{$^4$Shenzhen Institutes of Advanced Technology, Chinese Academy of Sciences}\\
{\{junbo.zhang, yuzheng\}@microsoft.com}, dekangqi@outlook.com
}
\maketitle
\begin{abstract}
Forecasting the flow of crowds is of great importance to traffic management and public safety, and very challenging as it is affected by many complex factors, such as inter-region traffic, events, and weather. We propose a deep-learning-based approach, called ST-ResNet, to \textit{collectively} forecast the inflow and outflow of crowds in each and every region of a city. We design an end-to-end structure of ST-ResNet based on unique properties of spatio-temporal data. More specifically, we employ the residual neural network framework to model the temporal closeness, period, and trend properties of crowd traffic. For each property, we design a branch of residual convolutional units, each of which models the spatial properties of crowd traffic. ST-ResNet learns to dynamically aggregate the output of the three residual neural networks based on data, assigning different weights to different branches and regions. The aggregation is further combined with external factors, such as weather and day of the week, to predict the final traffic of crowds in each and every region. 
Experiments on two types of crowd flows in Beijing and New York City (NYC) demonstrate that the proposed ST-ResNet outperforms six well-known methods. 
\end{abstract}
\section{Introduction}
Predicting crowd flows in a city is of great importance to traffic management and public safety \cite{Zheng2014AToISaTT}. For instance, massive crowds of  people streamed into a strip region at the 2015 New Year's Eve celebrations in Shanghai, resulting in a catastrophic stampede that killed  36 people. In mid-July of 2016, hundreds of ``Pokemon Go'' players ran through New York City's Central Park in hopes of catching a particularly rare digital monster, leading to a dangerous stampede there. If one can predict the crowd flow in a region, such tragedies can be mitigated or prevented by utilizing emergency mechanisms, such as conducting traffic control, sending out warnings, or evacuating people, in advance. 

\begin{figure}[!htbp]
\centering
\subfigure[{Inflow and outflow}]{\label{fig:in_out_flows}\includegraphics[width=.4\linewidth]{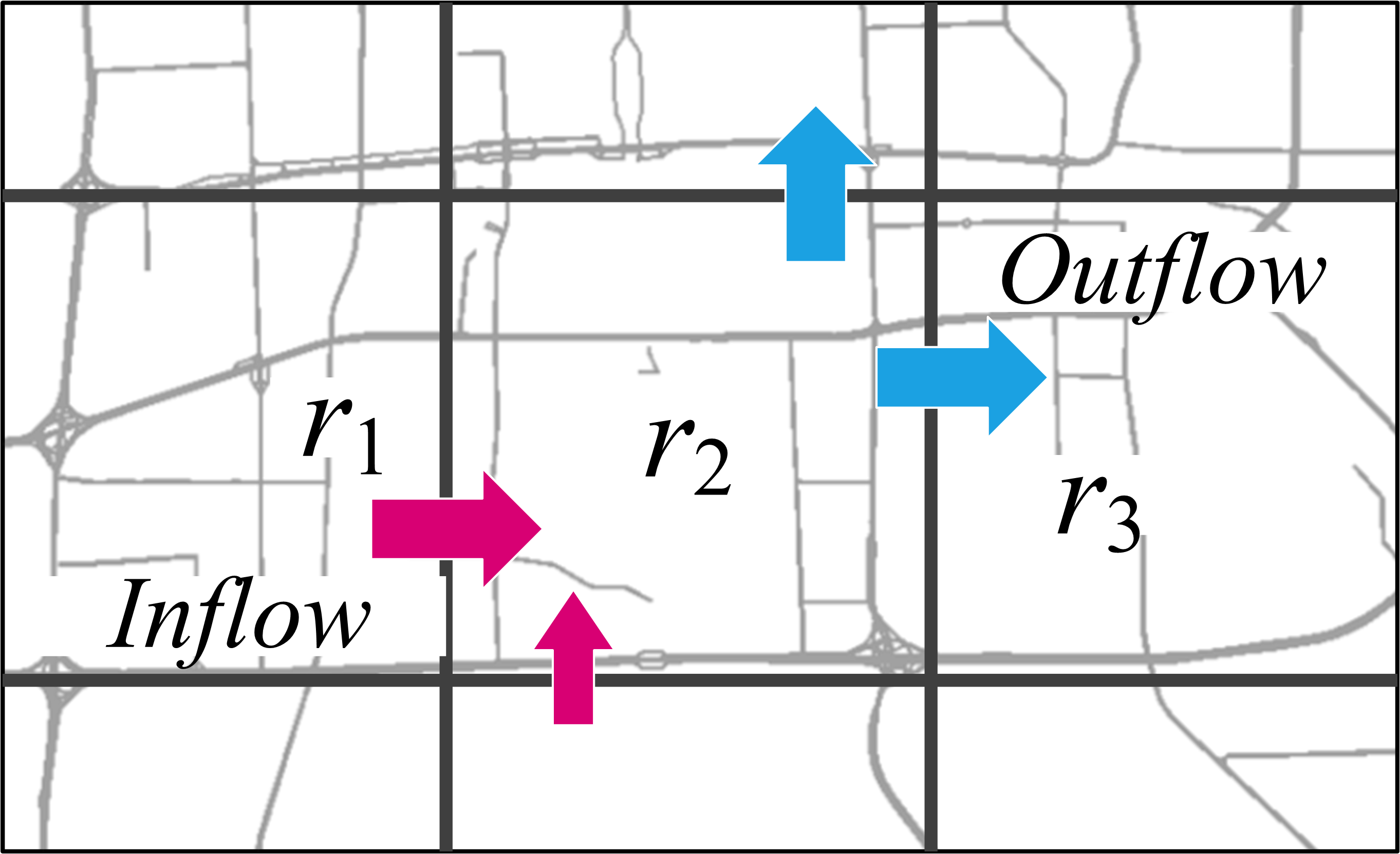}}
\hspace{1em}
\subfigure [Measurement of flows]{\label{fig:calc}\includegraphics[width=.4\linewidth]{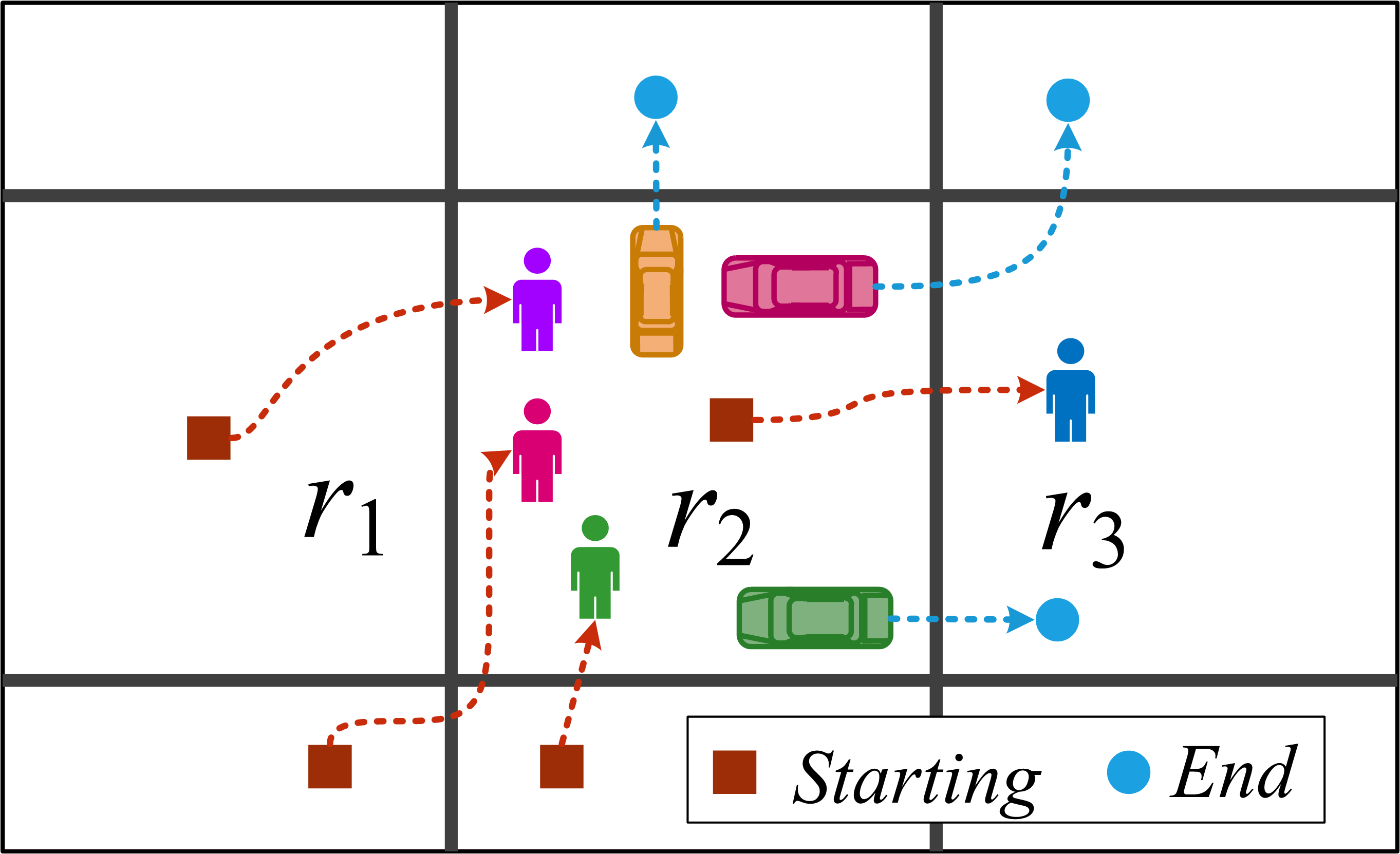}}
\caption{Crowd flows in a region}
\end{figure}

In this paper, we predict two types of crowd flows \cite{Zhang2016}: inflow and outflow, as shown in Figure~\ref{fig:in_out_flows}. 
Inflow is the total traffic of crowds entering a region from other places during a given time interval. 
Outflow denotes the total traffic of crowds leaving a region for other places during a given time interval. Both flows track the transition of crowds between regions. Knowing them is very beneficial for risk assessment and traffic management. 
Inflow/outflow can be measured by the number of pedestrians, the number of cars driven nearby roads, the number of people traveling on public transportation systems (\eg{}, metro, bus), or \textit{all of them together} if data is available. 
Figure~\ref{fig:calc} presents an example. We can use mobile phone signals to measure the number of pedestrians, showing that the inflow and outflow of $r_2$ are $(3, 1)$ respectively. Similarly, using the GPS trajectories of vehicles, two types of flows are $(0,3)$ respectively. 

\textit{Simultaneously} forecasting the inflow and outflow of crowds in each region of a city, however, is very challenging, affected by the following three complex factors: 
\begin{enumerate}
\item Spatial dependencies. The inflow of Region $r_2$ (shown in Figure~\ref{fig:in_out_flows}) is affected by outflows of nearby regions (like $r_1$) as well as distant regions. Likewise, the outflow of $r_2$ would affect inflows of other regions (\eg{}, $r_3$). The inflow of region $r_2$ would affect its own outflow as well. 
\item Temporal dependencies. The flow of crowds in a region is affected by recent time intervals, both near and far. For instance, a traffic congestion occurring at 8am will affect that of 9am. In addition, traffic conditions during morning rush hours may be similar on consecutive workdays, repeating every 24 hours. Furthermore, morning rush hours may gradually happen later as winter comes. When the temperature gradually drops and the sun rises later in the day, people get up later and later. 
\item External influence. Some external factors, such as weather conditions and events may change the flow of crowds tremendously in different regions of a city. 
\end{enumerate}

To tackle these challenges, we propose a deep spatio-temporal residual network (ST-ResNet) to \textit{collectively} predict inflow and outflow of crowds in every region. Our contributions are four-fold:
\begin{itemize}
\item ST-ResNet employs convolution-based residual networks to model nearby and distant spatial dependencies between any two regions in a city, while ensuring the model's prediction accuracy is not comprised by the deep structure of the neural network. 
\item We summarize the temporal properties of crowd flows into three categories, consisting of temporal closeness, period, and trend. ST-ResNet uses three residual networks to model these properties, respectively. 
\item ST-ResNet dynamically aggregates the output of the three aforementioned networks, assigning different weights to different branches and regions. The aggregation is further combined with external factors (\eg{}, weather). 
\item We evaluate our approach using Beijing taxicabs' trajectories and meteorological data, and NYC bike trajectory data. The results demonstrate the advantages of our approach compared with 6 baselines. 
\end{itemize}

\section{Preliminaries}
In this section, we briefly revisit the crowd flows prediction problem \cite{Zhang2016,Hoang2016} and introduce deep residual learning \cite{He2016apa}. 
\subsection{Formulation of Crowd Flows Problem}
\begin{figure}[!htbp]
\centering
\label{fig:flows}\includegraphics[width=.9\linewidth]{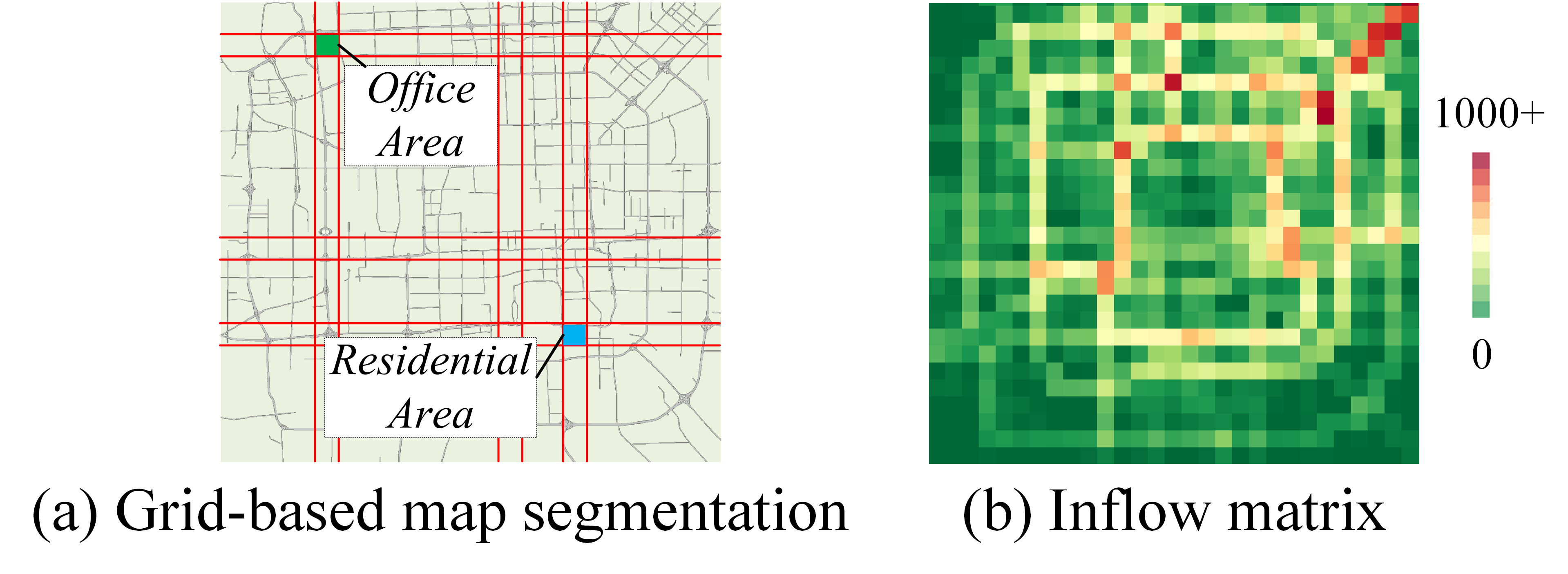}
\caption{Regions in Beijing: (a) Grid-based map segmentation; (b) inflows in every region of Beijing}
\label{fig:map}
\end{figure}

\begin{definition}[Region \cite{Zhang2016}]\label{def:region}
There are many definitions of a location in terms of different granularities and semantic meanings. In this study, we partition a city into an $I \times J$ grid map based on the longitude and latitude where a grid denotes a region, as shown in Figure~\ref{fig:map}(a).
\end{definition}

\begin{definition}[Inflow/outflow \cite{Zhang2016}]\label{def:flow}
Let $\mathbb P$ be a collection of trajectories at the $t^{th}$ time interval. 
For a grid $(i, j)$ that lies at the $i^{th}$ row and the $j^{th}$ column, the inflow and outflow of the crowds at the time interval $t$ are defined respectively as 
\begin{eqnarray}
x_t^{in, i, j} &=& \sum\limits_{Tr \in \mathbb P} |\{k>1 | g_{k-1} \not\in (i, j) \wedge g_k \in (i, j)\}| \nonumber \\
x_t^{out, i, j} &=& \sum\limits_{Tr \in \mathbb P} |\{k\geq 1 | g_k \in (i, j) \wedge g_{k+1} \not\in (i, j)\}| \nonumber
\end{eqnarray}
where $Tr: g_1 \rightarrow g_2 \rightarrow \cdots  \rightarrow g_{|Tr|}$ is a trajectory in $\mathbb P$, and
$g_k$ is the geospatial coordinate; $g_k \in (i, j)$ means the point $g_k$ lies within grid $(i, j)$, and vice versa; $|\cdot|$ denotes the cardinality of a set. 
\end{definition}
At the $t^{th}$ time interval, inflow and outflow in all $I \times J$ regions can be denoted as a tensor 
$\mathbf X_t \in \mathbb R ^{2\times I\times J}$ where $(\mathbf X_t)_{0,i, j}=x_t^{in, i, j}$, $(\mathbf X_t)_{1,i, j}=x_t^{out, i, j}$. 
The inflow matrix is shown in Figure~\ref{fig:map}(b). 

Formally, for a dynamical system over a spatial region represented by a $I \times J$ grid map, there are 2 types of flows in each grid over time. 
Thus, the observation at any time can be represented by a tensor $\mathbf X \in \mathbb R ^{2 \times I\times J}$. 
\begin{problem}
Given the historical observations $\{\mathbf X_t| t=0, \cdots, n-1\}$, predict $\mathbf X_n$.
\end{problem}

\subsection{Deep Residual Learning}
Deep residual learning \cite{He2015apa} allows convolution neural networks to have a super deep structure of 100 layers, even over-1000 layers. 
And this method has shown state-of-the-art results on multiple challenging recognition tasks, including image classification, object detection, segmentation and localization \cite{He2015apa}. 

Formally, a residual unit with an identity mapping \cite{He2016apa} is defined as:
\begin{equation}
\mathbf X^{(l+1)} = \mathbf X^{(l)} + \mathcal F(\mathbf X^{(l)})
\end{equation}
where $\mathbf X^{(l)}$ and $\mathbf X^{(l+1)}$ are the input and output of the $l^{th}$ residual unit, respectively; $\mathcal F$ is a residual function, \eg{}, a stack of two $3\times 3$ convolution layers in \cite{He2015apa}. 
The central idea of the residual learning is to learn the additive residual function $\mathcal F$ with respect to $\mathbf X^{(l)}$ \cite{He2016apa}. 

\begin{figure}[!b]
\centering
\includegraphics[width=0.95\linewidth]{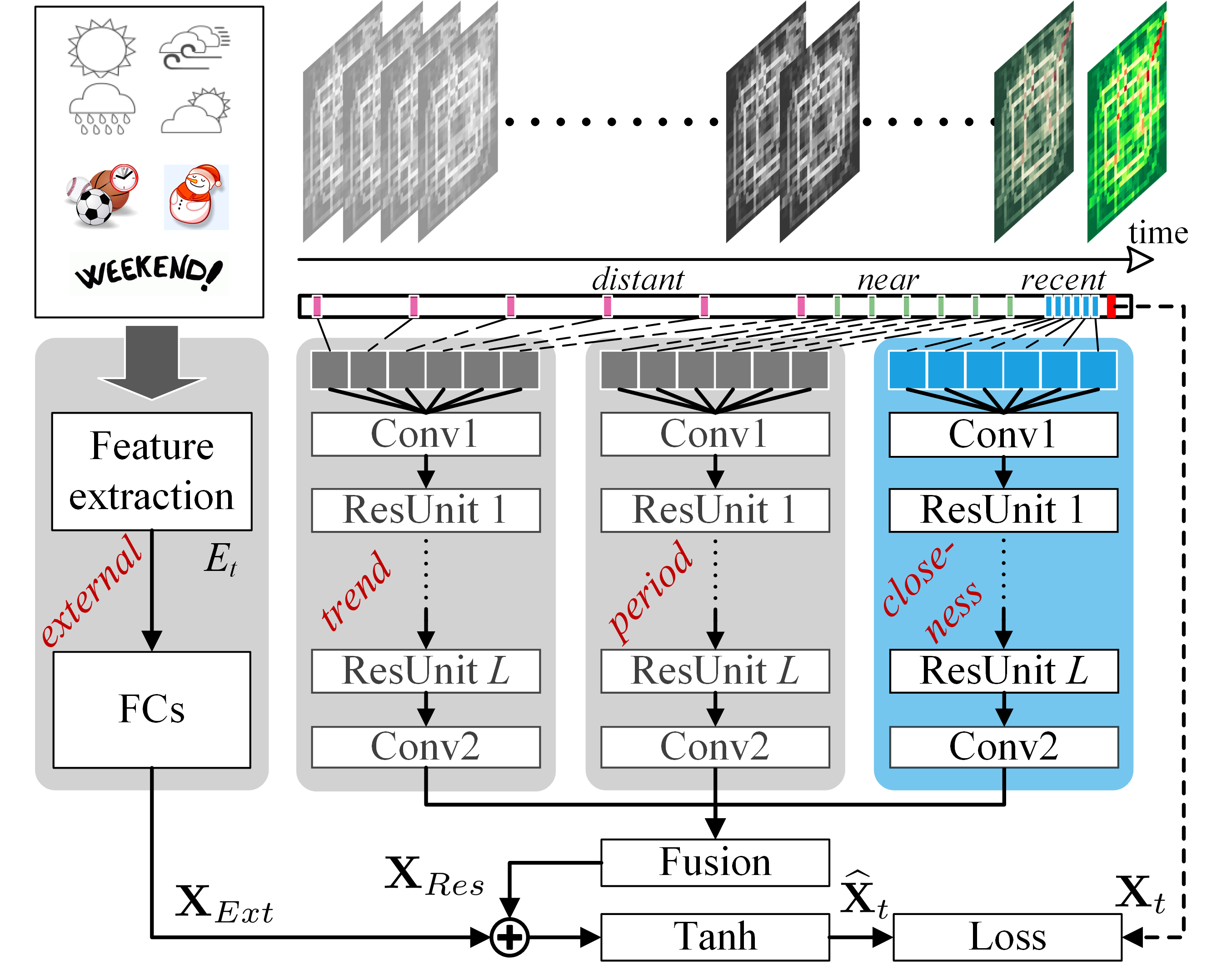}
\caption{ST-ResNet architecture. Conv: Convolution; \\ResUnit: Residual Unit; FC: Fully-connected.}
\label{fig:STResNet}
\end{figure}

\section{Deep Spatio-Temporal Residual Networks}
Figure~\ref{fig:STResNet} presents the architecture of ST-ResNet, which is comprised of four major components modeling temporal \textit{closeness}, \textit{period}, \textit{trend}, and \textit{external} influence, respectively. As illustrated in the top-right part of Figure~\ref{fig:STResNet}, we first turn Inflow and outflow throughout a city at each time interval into a 2-channel image-like matrix respectively, using the approach introduced in Definitions 1 and 2. We then divide the time axis into three fragments, denoting recent time, near history and distant history. The 2-channel flow matrices of intervals in each time fragment are then fed into the first three components separately to model the aforementioned three temporal properties: \textit{closeness}, \textit{period} and \textit{trend}, respectively. The first three components share the same network structure with a convolutional neural network followed by a Residual Unit sequence. Such structure captures the spatial dependency between nearby and distant regions. In the \textit{external} component, we manually extract some features from external datasets, such as weather conditions and events, feeding them into a two-layer fully-connected neural network. The outputs of the first three components are fused as $\mathbf X_{Res}$ based on parameter matrices, which assign different weights to the results of different components in different regions. $\mathbf X_{Res}$ is further integrated with the output of the external component $\mathbf X_{Ext}$. Finally, the aggregation is mapped into $[-1, 1]$ by a Tanh function, which yields a faster convergence than the standard logistic function in the process of back-propagation learning \cite{lecun2012efficient}. 

\subsection{Structures of the First Three Components}
\begin{figure}[!htbp]
\centering
\subfigure[Convolutions  ]{\label{fig:holiday}\includegraphics[width=.7\linewidth]{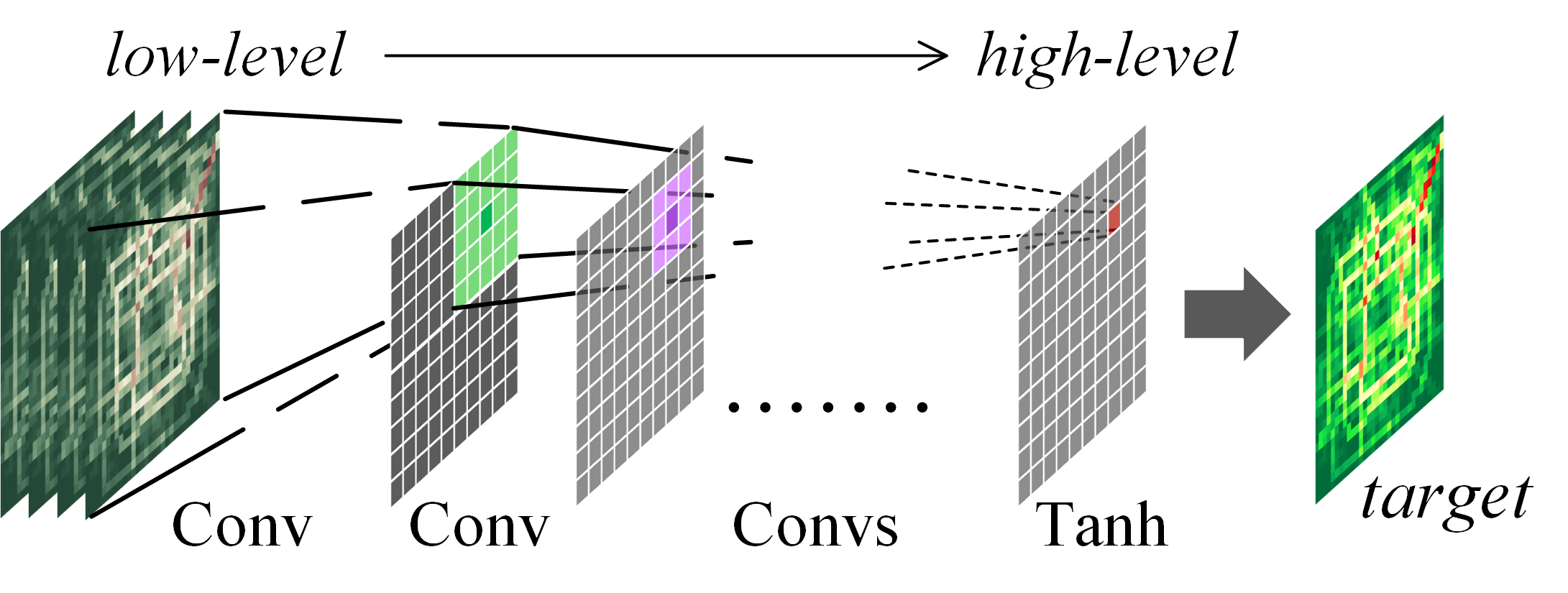}}
\subfigure[Residual Unit]
{\label{fig:weather}\includegraphics[width=.7\linewidth]{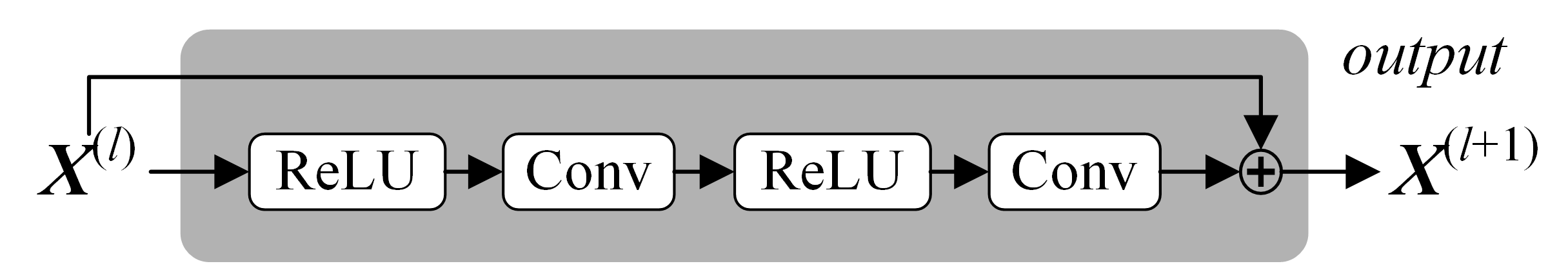}}
\caption{Convolution and residual unit}
\label{fig:ResUnit}
\end{figure}

The first three components (\ie{} \textit{closeness}, \textit{period}, \textit{trend}) share the same network structure, which is composed of two sub-components: convolution and residual unit, as shown in Figure~\ref{fig:ResUnit}.

\noindent\textit{\textbf{Convolution.}} 
A city usually has a very large size, containing many regions with different distances. 
Intuitively, the flow of crowds in nearby regions may affect each other, which can be effectively handled by the convolutional neural network (CNN) that has shown its powerful ability to hierarchically capture the spatial structural information \cite{LeCun1998PotI}. 
In addition, subway systems and highways connect two locations with a far distance, leading to the dependency between distant regions. In order to capture the spatial dependency of any region, we need to design a CNN with many layers because one convolution only accounts for spatial near dependencies, limited by the size of their kernels. 
The same problem also has been found in the video sequence generating task where the input and output have the same resolution \cite{Mathieu2015apa}. 
Several methods have been introduced to avoid the loss of resolution brought about by subsampling while preserving distant dependencies \cite{Long2015}. Being different from the classical CNN, we do not use subsampling, but only convolutions \cite{Jain2007}. As shown in Figure~\ref{fig:ResUnit}(a), there are three multiple levels of feature maps that are connected with a few convolutions. 
We find that a node in the high-level feature map depends on nine nodes of the middle-level feature map, those of which depend on all nodes in the lower-level feature map (\ie{} input). It means one convolution naturally captures spatial near dependencies, and a stack of convolutions can further capture distant even citywide dependencies. 

The \textit{closeness} component of Figure~\ref{fig:STResNet} adopts a few 2-channel flows matrices of intervals in the recent time to model temporal \textit{closeness} dependence. 
Let the recent fragment be $[\mathbf X_{t-{l_c}}, \mathbf X_{t-{(l_c -1)}} ,\cdots, \mathbf X_{t-1}]$, which is also known as the \textit{closeness} dependent sequence. 
We first concatenate them along with the first axis (\ie{} time interval) as one tensor $\mathbf X_c^{(0)} \in \mathbb R^{2l_c \times I \times J}$, which is followed by a convolution (\ie{} \texttt{Conv1} shown in Figure~\ref{fig:STResNet}) as:
\begin{equation}\label{eq:conv}
\mathbf X_{c}^{(1)} = f \left(W^{(1)}_{c} *\mathbf X_c^{(0)}+ b^{(1)}_{c} \right) \nonumber
\end{equation}
where $*$ denotes the convolution\footnote{To make the input and output have the same size (\ie{} $I \times J$) in a convolutional operator, we employ a border-mode which allows a filter to go outside the border of an input, padding each area outside the border with a zero. }; $f$ is an activation function, \textit{e.g.} the rectifier $f(z) := \max (0,z)$ \cite{Krizhevsky2012}; $W^{(1)}_{c}, b_c^{(1)}$ are the learnable parameters in the first layer.  

\noindent\textit{\textbf{Residual Unit.}} 
It is a well-known fact that very deep convolutional networks compromise training effectiveness though the well-known activation function (\eg{} ReLU) and regularization techniques are applied \cite{Ioffe2015,Krizhevsky2012,Nair2010}. 
On the other hand, we still need a very deep network to capture very large citywide dependencies. 
For a typical crowd flows data, assume that the input size is $32 \times 32$, and the kernel size of convolution is fixed to $3 \times 3$, if we want to model citywide dependencies (\ie{}, each node in high-level layer depends on all nodes of the input), it needs more than 15 consecutive convolutional layers. 
To address this issue, we employ residual learning \cite{He2015apa} in our model, which have been demonstrated to be very effective for training super deep neural networks of over-1000 layers. 

In our ST-ResNet (see Figure~\ref{fig:STResNet}), we stack $L$ residual units upon \texttt{Conv1} as follows, 
\begin{equation}\label{eq:res}
\mathbf X_c^{(l+1)} = \mathbf X_c^{(l)} + \mathcal F(\mathbf X_c^{(l)}; \theta_c^{(l)}), l=1,\cdots, L 
\end{equation}
where $\mathcal F$ is the residual function (\ie{} two combinations of ``ReLU + Convolution'', see Figure~\ref{fig:ResUnit}(b)), and $\theta^{(l)}$ includes all learnable parameters in the $l^{th}$ residual unit. We also attempt \textit{Batch Normalization} (BN) \cite{Ioffe2015} that is added before ReLU. 
On top of the $L^{th}$ residual unit, 
we append a convolutional layer (\ie{} \texttt{Conv2} shown in Figure~\ref{fig:STResNet}). 
With 2 convolutions and $L$ residual units, the output of the \textit{closeness} component of Figure~\ref{fig:STResNet} is $\mathbf X_c^{(L+2)}$. 

Likewise, using the above operations, we can construct the \textit{period} and \textit{trend} components of Figure~\ref{fig:STResNet}. 
Assume that there are $l_p$ time intervals from the period fragment and the period is $p$. Therefore, the \textit{period} dependent sequence is $[\mathbf X_{t-{l_p} \cdot p}, \mathbf X_{t-({l_p}-1)\cdot p}, \cdots, \mathbf X_{t-p}]$. 
With the convolutional operation and $L$ residual units like in Eqs.~\ref{eq:conv} and \ref{eq:res}, the output of the \textit{period} component is $\mathbf X_p^{(L+2)}$. 
Meanwhile, the output of the \textit{trend} component is $\mathbf X_q^{(L+2)}$ with the input $[\mathbf X_{t-{l_q} \cdot q}, \mathbf X_{t-({l_q}-1)\cdot q}, \cdots, \mathbf X_{t-q}]$ where $l_q$ is the length of the \textit{trend} dependent sequence and $q$ is the trend span. 
Note that $p$ and $q$ are actually two different types of periods. In the detailed implementation, $p$ is equal to one-day that describes daily periodicity, and $q$ is equal to one-week that reveals the weekly trend. 

\subsection{The Structure of the External Component}
Traffic flows can be affected by many complex external factors, such as weather and event. 
Figure~\ref{fig:holiday} shows that crowd flows during holidays (Chinese Spring Festival) can be significantly different from the flows during normal days. Figure~\ref{fig:weather} shows that heavy rain sharply reduces the crowd flows at Office Area compared to the same day of the latter week. 
Let $E_t$ be the feature vector that represents these external factors at predicted time interval $t$. 
In our implementation, we mainly consider weather, holiday event, and metadata (\ie{} DayOfWeek, Weekday/Weekend). 
The details are introduced in Table~\ref{tab:datasets}. 
To predict flows at time interval $t$, the holiday event and metadata can be directly obtained. However, the weather at future time interval $t$ is unknown. Instead, one can use the forecasting weather at time interval $t$ or the approximate weather at time interval $t-1$. 
Formally, we stack two fully-connected layers upon $E_t$, the first layer can be viewed as an embedding layer for each sub-factor followed by an activation. The second layer is used to map low to high dimensions that have the same shape as $\mathbf X_t$. The output of the 
\textit{external} component of Figure~\ref{fig:STResNet} is denoted as $\mathbf X_{Ext}$ with the parameters $\theta_{Ext}$. 

\begin{figure}[!htbp]
\centering
\subfigure[{\small Feb 8-14 (red), Feb 15-21 (green), 2016}]{\label{fig:holiday}\includegraphics[width=.45\linewidth]{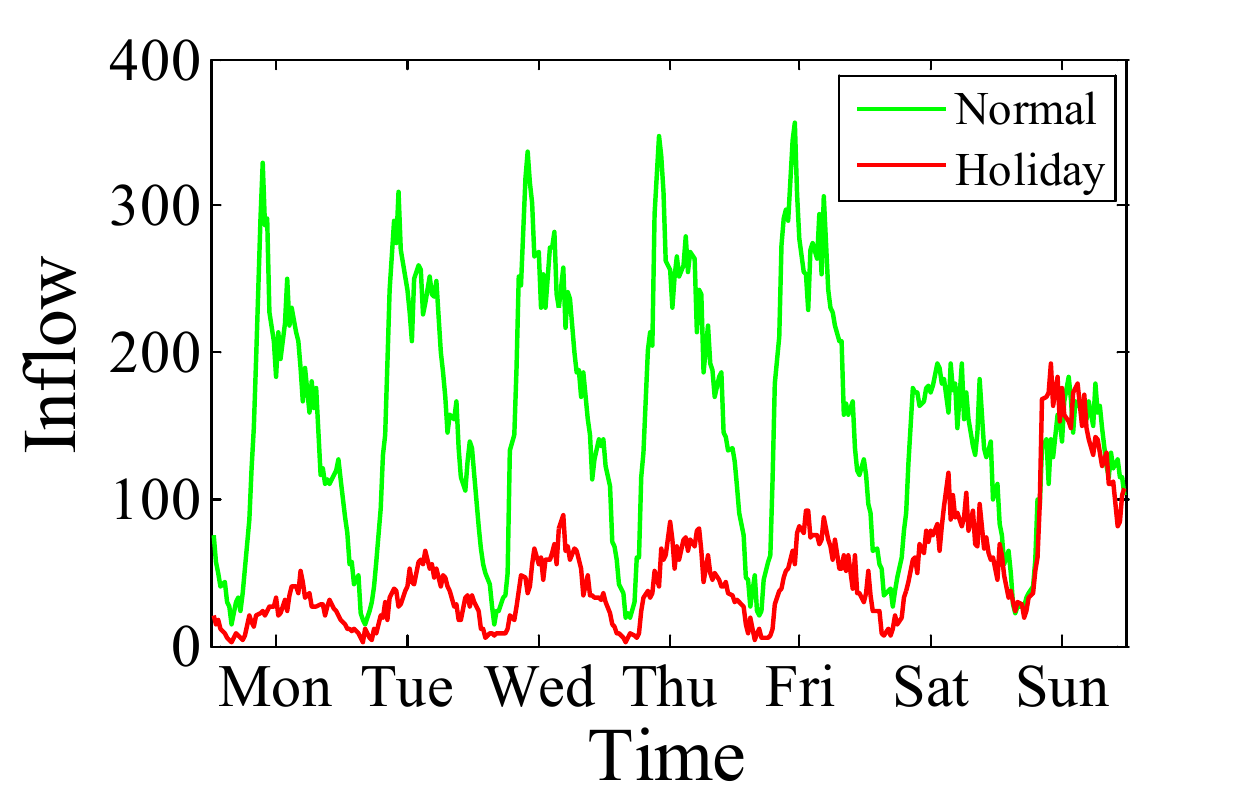}}
\hspace{1em}
\subfigure[{\small Aug 10-12 (red), Aug 17-19 (green), 2013}]
{\label{fig:weather}\includegraphics[width=.45\linewidth]{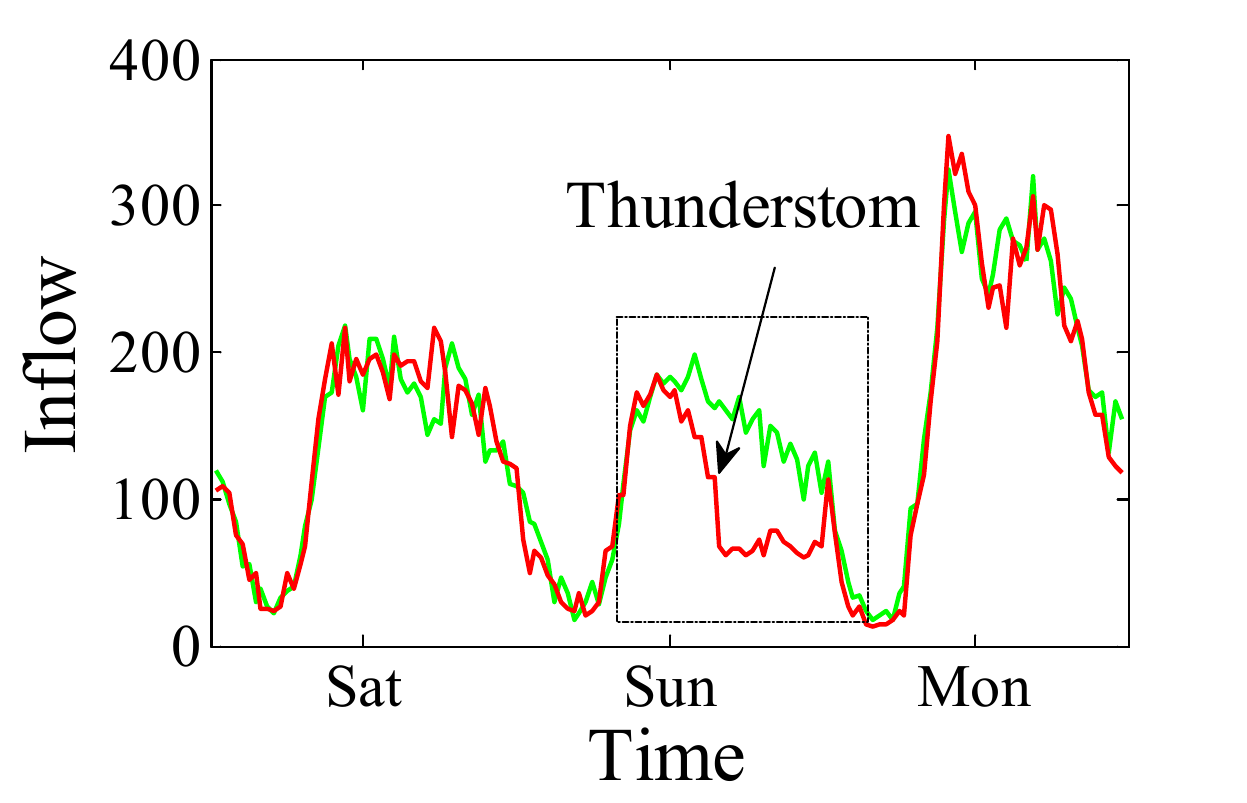}}
\caption{Effects of holidays and weather in Office Area of Beijing (the region is shown in Figure~\ref{fig:map}(a)). }
\label{fig:external}
\end{figure}

\subsection{Fusion}
In this section, we discuss how to fuse four components of Figure~\ref{fig:STResNet}. We first fuse the first three components with a parametric-matrix-based fusion method, which is then further combined with the \textit{external} component. 

Figures~\ref{fig:temporal}(a) and (d) show the ratio curves using Beijing trajectory data presented in Table~\ref{tab:datasets} where $x$-axis is time gap between two time intervals and $y$-axis is the average ratio value between arbitrary two inflows that have the same time gap. 
The curves from two different regions all show an empirical temporal correlation in time series, namely, inflows of recent time intervals are more relevant than ones of distant time intervals, which implies temporal \textit{closeness}. 
The two curves have different shapes, which demonstrates that different regions may have different characteristics of closeness. 
Figures~\ref{fig:temporal}(b) and (e) depict inflows at all time intervals of 7 days. We can see the obvious \textit{daily periodicity} in both regions. In Office Area, the peak values on weekdays are much higher than ones on weekends. Residential Area has similar peak values for both weekdays and weekends. 
Figures~\ref{fig:temporal}(c) and (f) describe inflows at a certain time interval (9:00pm-9:30pm) of Tuesday from March 2015 and June 2015. 
As time goes by, the inflow progressively decreases in Office Area, and increases in Residential Area. 
It shows the different trends in different regions.  
In summary, inflows of two regions are all affected by \textit{closeness}, \textit{period}, and \textit{trend}, but the degrees of influence may be very different. 
We also find the same properties in other regions as well as their outflows. 

\begin{figure}[!htbp]
\centering\label{fig:temporal}\includegraphics[width=1.\linewidth]{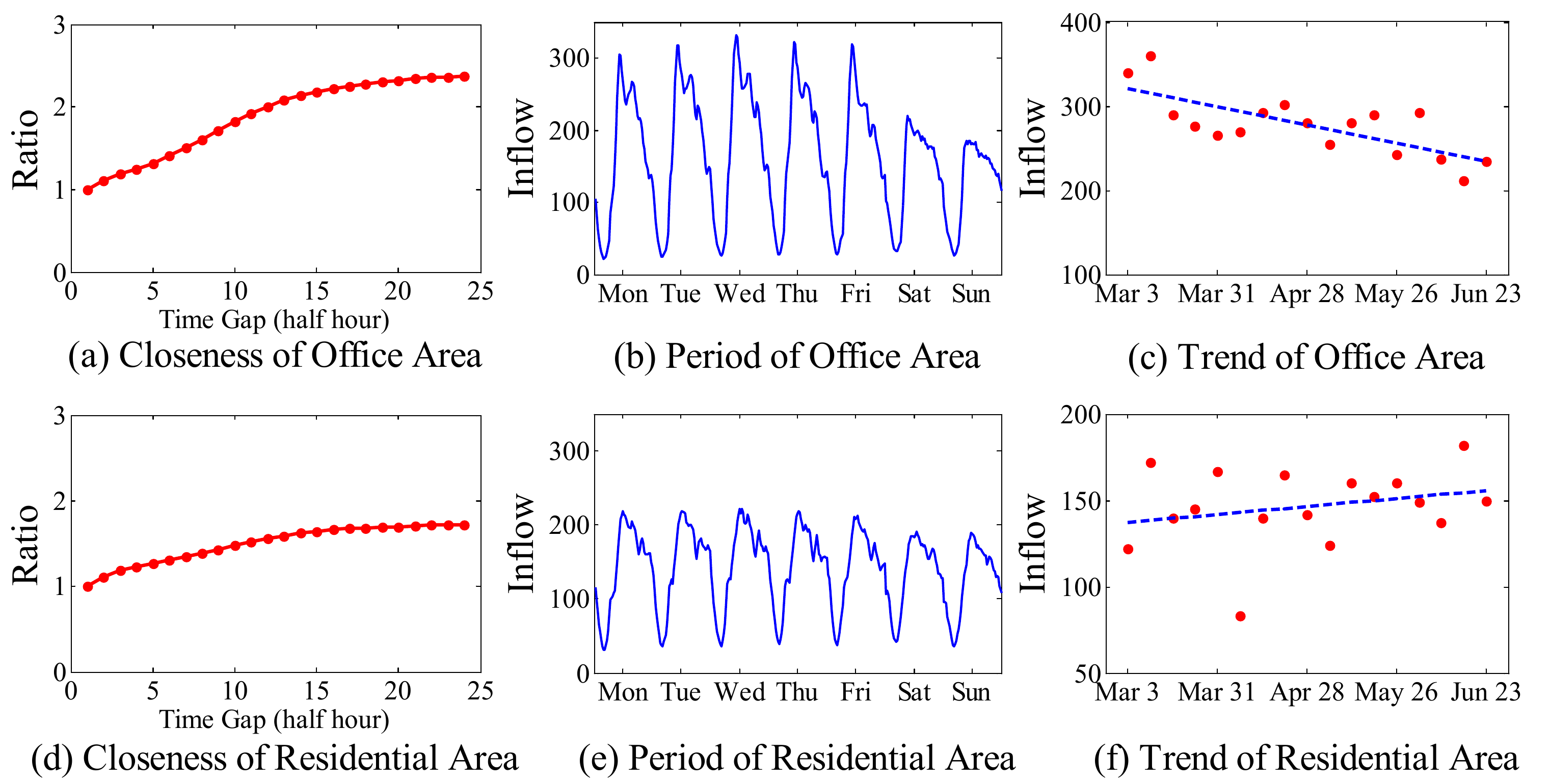}
\caption{Temporal dependencies (Office Area and Residential Area are shown in Figure~\ref{fig:map}(a))}
\label{fig:temporal}
\end{figure}

Above all, the different regions are all affected by \textit{closeness}, \textit{period} and \textit{trend}, but the degrees of influence may be different. 
Inspired by these observations, we propose a parametric-matrix-based fusion method. 

\noindent{\textbf{\textit{Parametric-matrix-based fusion}}}. 
We fuse the first three components (\ie{} \textit{closeness}, \textit{period}, \textit{trend}) of Figure~\ref{fig:STResNet} as follows
\begin{equation}\label{eq:res_out}
\mathbf X_{Res}=\mathbf W_c \circ \mathbf X_c^{(L+2)} + \mathbf W_p \circ \mathbf X_p^{(L+2)} + \mathbf W_q \circ \mathbf X_q^{(L+2)}
\end{equation}
where $\circ$ is Hadamard product (\ie, element-wise multiplication), $\mathbf W_c$, $\mathbf W_p$ and $\mathbf W_q$ are the learnable parameters that adjust the degrees affected by closeness, period and trend, respectively. 

\noindent{\textbf{\textit{Fusing the external component}}}. We here directly merge the output of the first three components with that of the \textit{external} component, as shown in Figure~\ref{fig:STResNet}. Finally, the predicted value at the $t^{th}$ time interval, denoted by $\widehat{\mathbf X}_t$, is defined as 
\begin{equation}\label{eq:output}
\widehat{\mathbf X}_t = \tanh (\mathbf X_{Res} + \mathbf X_{Ext})
\end{equation}
where $\tanh$ is a hyperbolic tangent that ensures the output values are between -1 and 1.

Our ST-ResNet can be trained to predict $\mathbf X_t$ from three sequences of flow matrices and external factor features by minimizing mean squared error between the predicted flow matrix and the true flow matrix: 

\begin{equation} \label{eq:loss}
\mathcal L(\theta) = \|\mathbf X_t - \widehat{\mathbf X}_t \| ^2_2
\end{equation} 
where $\theta$ are all learnable parameters in the ST-ResNet. 

\subsection{Algorithm and Optimization}
Algorithm~\ref{Alg:STResNet} outlines the ST-ResNet training process. 
We first construct the training instances from the original sequence data (lines 1-6). 
Then, ST-ResNet is trained via backpropagation and Adam \cite{Kingma2014apa} (lines 7-11). 

\begin{algorithm}[!htbp]\label{Alg:STResNet}\fontsize{9}{9}\selectfont
\KwIn{Historical observations: $\{\mathbf X_0, \cdots, \mathbf X_{n-1}\}$; \\
\qquad\quad external features: $\{E_0, \cdots, E_{n-1}\}$; \\
\qquad\quad lengths of \textit{closeness}, \textit{period}, \textit{trend} sequences: $l_c$, $l_p, l_q;$\\
\qquad\quad peroid: $p$; trend span: $q$.}
\KwOut{Learned ST-ResNet model}
{	\DontPrintSemicolon
	{\tcp*[h]{construct training instances}}\;
	$\mathcal D \longleftarrow \emptyset$\; 
	\For{all available time interval $t (1\leq t \leq n-1)$}
	{
	 	$\mathcal S_c=[\mathbf X_{t-{l_c}}, \mathbf X_{t-({l_c}-1)}, \cdots, \mathbf X_{t-1}]$ \;
	 	$\mathcal S_p=[\mathbf X_{t-{l_p} \cdot p}, \mathbf X_{t-({l_p}-1)\cdot p}, \cdots, \mathbf X_{t-p}]$ \;
	 	$\mathcal S_q=[\mathbf X_{t-{l_q} \cdot q}, \mathbf X_{t-({l_q}-1)\cdot q}, \cdots, \mathbf X_{t-q}]$ \;
	 	\tcp*[h]{$\mathbf X_t$ is the target at time $t$}\;
	 	put an training instance $(\{\mathcal S_c, \mathcal S_p,\mathcal S_q, E_t\},  \mathbf X_t)$ into  $\mathcal D$\;
	}
	{\tcp*[h]{train the model}}\;
    initialize all learnable parameters $\theta$ in ST-ResNet\;
	\Repeat{stopping criteria is met}
	{
	randomly select a batch of instances $\mathcal D_b$ from $\mathcal D$\;
	find $\theta$ by minimizing the objective (\ref{eq:loss}) with $\mathcal D_b$\;
	}
}
\caption{ST-ResNet Training Algorithm}
\end{algorithm}

\section{Experiments}
\subsection{Settings}
\noindent\textbf{Datasets.} We use two different sets of data as shown in Table~\ref{tab:datasets}. Each dataset contains two sub-datasets: trajectories and weather, as detailed as follows. 

\begin{itemize}
\item \textbf{TaxiBJ}: Trajectoriy data is the taxicab GPS data and meteorology data in Beijing from four time intervals: 1st Jul. 2013 - 30th Otc. 2013, 1st Mar. 2014 - 30th Jun. 2014, 1st Mar. 2015 - 30th Jun. 2015, 1st Nov. 2015 - 10th Apr. 2016. 
Using Definition~\ref{def:flow}, we obtain two types of crowd flows. 
We choose data from the last four weeks as the testing data, and all data before that as training data. 
\item \textbf{BikeNYC}: Trajectory data is taken from the NYC Bike system in 2014, from Apr. 1st to Sept. 30th. Trip data includes: trip duration, starting and ending station IDs, and start and end times. Among the data, the last 10 days are chosen as testing data, and the others as training data. 
\end{itemize}

\begin{table}[!htbp]\fontsize{9}{9}\selectfont
\tabcolsep 0pt 
\begin{center}
\caption{Datasets (holidays include adjacent weekends).}\label{tab:datasets}
\begin{tabular}{ccc}
\hline
\textbf{Dataset} & \textbf{TaxiBJ} & \textbf{BikeNYC} \\
\hline
Data type & Taxi GPS & Bike rent\\
Location & Beijing & New York \\
\cline{2-3}
\multirow{4}*{Time Span} & 7/1/2013 - 10/30/2013 & \\
&3/1/2014 - 6/30/2014&4/1/2014 - \\
&3/1/2015 - 6/30/2015&9/30/2014\\
&11/1/2015 - 4/10/2016&\\
\cline{2-3}
Time interval & 30 minutes & 1 hour \\
Gird map size & (32, 32) & (16, 8) \\
\hline
\multicolumn{3}{c}{\textbf{Trajectory data}} \\
Average sampling rate (s) & $\sim$ 60 & $\setminus$ \\
\# taxis/bikes & 34,000+ & 6,800+ \\
\# available time interval & 22,459  & 4,392 \\
\hline
\multicolumn{3}{c}{\textbf{External factors (holidays and meteorology)}} \\
\# holidays & 41 & 20 \\
Weather conditions & 16 types (\eg{}, Sunny, Rainy) & $\setminus$ \\
Temperature / $^\circ$C  & $[-24.6, 41.0]$ & $\setminus$ \\ 
Wind speed / mph & $[0, 48.6]$ & $\setminus$ \\
\hline
\end{tabular}
\end{center}
\end{table}

\noindent\textbf{Baselines}. We compare our ST-ResNet with the following 6 baselines: 
\begin{itemize}
\item \textbf{HA}: We predict inflow and outflow of crowds by the average value of historical inflow and outflow in the corresponding periods, \eg{}, 9:00am-9:30am on Tuesday, its corresponding periods are all historical time intervals from 9:00am to 9:30am on all historical Tuesdays. 
\item \textbf{ARIMA}: Auto-Regressive Integrated Moving Average (ARIMA) is a well-known model for understanding and predicting future values in a time series. 
\item \textbf{SARIMA}: Seasonal ARIMA. 
\item \textbf{VAR}: Vector Auto-Regressive (VAR) is a more advanced spatio-temporal model, which can capture the pairwise relationships among all flows, and has heavy computational costs due to the large number of parameters. 
\item \textbf{ST-ANN}: It first extracts spatial (nearby 8 regions' values) and temporal (8 previous time intervals) features, then fed into an artificial neural network. 
\item \textbf{DeepST} \cite{Zhang2016}: a deep neural network (DNN)-based prediction model for spatio-temporal data, which shows state-of-the-art results on crowd flows prediction. It has 4 variants, including DeepST-C, DeepST-CP, DeepST-CPT, and DeepST-CPTM, which focus on different temporal dependencies and external factors. 
\end{itemize}

\noindent\textbf{Preprocessing.} In the output of the ST-ResNet, we use $\tanh$ as our final activation (see Eq.~\ref{eq:output}), whose range is between -1 and 1. Here, we use the Min-Max normalization method to scale the data into the range $[-1, 1]$. In the evaluation, we re-scale the predicted value back to the normal values, compared with the groundtruth. For external factors, we use one-hot coding to transform metadata (\ie{}, DayOfWeek, Weekend/Weekday), holidays and weather conditions into binary vectors, and use Min-Max normalization to scale the Temperature and Wind speed into the range $[0, 1]$. 

\noindent\textbf{Hyperparameters.} The python libraries, including Theano \cite{TDT2016ae} and Keras \cite{Chollet2015}, are used to build our models. 
The convolutions of \texttt{Conv1} and all residual units use 64 filters of size $3\times 3$, and \texttt{Conv2} uses a convolution with 2 filters of size  $3\times 3$. 
The batch size is 32. We select 90\% of the training data for training each model, and the remaining 10\% is chosen as the validation set, which is used to early-stop our training algorithm for each model based on the best validation score. 
Afterwards, we continue to train the model on the full training data for a fixed number of epochs (\eg{}, 10, 100 epochs). 
There are 5 extra hyperparamers in our ST-ResNet, of which $p$ and $q$ are empirically fixed to one-day and one-week, respectively. 
For lengths of the three dependent sequences, we set them as: $l_c \in \{3,4,5\}, l_p \in \{1,2,3,4\}, l_q \in \{1,2,3,4\}$. 

\noindent\textbf{Evaluation Metric}: We measure our method by Root Mean Square Error (RMSE) as
\begin{equation}
RMSE = \sqrt { \frac{1}{z} \sum_i (x_i - \hat x_i)^2 } \nonumber
\end{equation}
where $\hat x$ and $x$ are the predicted value and ground thuth, respectively; $z$ is the number of all predicted values. 
\subsection{Results on TaxiBJ}
We first give the comparison with 6 other models on TaxiBJ, as shown in Table~\ref{tab:TaxiBJ}. 
We give 7 variants of ST-ResNet with different layers and different factors. Taking L12-E for example, it considers all available external factors and has 12 residual units, each of which is comprised of two convolutional layers. 
We observe that all of these 7 models are better than 6 baselines. Comparing with the previous state-of-the-art models, L12-E-BN reduces error to $16.69$, which significantly improves accuracy. 

\begin{table}[!htbp]\fontsize{9}{9}\selectfont
\tabcolsep 3pt
\begin{center}
\caption{Comparison among different methods on TaxiBJ}
\label{tab:TaxiBJ}
\begin{tabular}{ll|c}
\hline
Model & & RMSE \\
\hline
HA	&& 57.69\\
ARIMA	&& 22.78\\
SARIMA &&	26.88\\
VAR &	&22.88\\
ST-ANN &&	19.57\\
DeepST &&	18.18\\
\hline
&{\textbf{ST-ResNet} [ours]}  & \\
L2-E & 2 residual units + E	&17.67\\
L4-E & 4 residual units + E	&17.51\\
L12-E & 12 residual units + E	&16.89\\
L12-E-BN & L12-E with BN	&\textbf{16.69}\\
L12-single-E	& 12 residual units (1 conv) + E &17.40\\
L12 & 12 residual units &17.00 \\
L12-E-noFusion & 12 residual units + E without fusion  & 17.96\\
\hline
\end{tabular}
\end{center}
\end{table}

\noindent\textbf{Effects of Different Components}. Let \textbf{L12-E} be the compared model. 
\begin{itemize}
\item \textit{Number of residual units}: Results of L2-E, L4-E and L12-E show that RMSE decreases as the number of residual units increases. Using residual learning, the deeper the network is, the more accurate the results will be. 
\item \textit{Internal structure of residual unit}: We attempt three different types of residual units. L12-E adopts the standard \textit{Residual Unit} (see Figure~\ref{fig:ResUnit}(b)). Compared with L12-E, \textit{Residual Unit} of L12-single-E only contains 1 ReLU followed by 1 convolution, and  \textit{Residual Unit} of L12-E-BN added two batch normalization layers, each of which is inserted before ReLU. We observe that L12-single-E is worse than L12-E, and L12-E-BN is the best, demonstrating the effectiveness of \textit{batch normalization}. 
\item \textit{External factors}: L12-E considers the external factors, including meteorology data, holiday events and metadata. If not, the model is degraded as L12. The results indicate that L12-E is better than L12, pointing out that external factors are always beneficial. 
\item \textit{Parametric-matrix-based fusion}: Being different with L12-E, L12-E-noFusion donot use parametric-matrix-based fusion (see Eq.~\ref{eq:res_out}). Instead, L12-E-noFusion use a straightforward method for fusing, \ie{}, $\mathbf X_c^{(L+2)} + \mathbf X_p^{(L+2)} + \mathbf X_q^{(L+2)}$. It shows the error greatly increases, which demonstrates the effectiveness of our proposed parametric-matrix-based fusion. 
\end{itemize}

\subsection{Results on BikeNYC}
Table~\ref{tab:BikeNYC} shows the results of our model and other baselines on BikeNYC. Being different from TaxiBJ, BikeNYC consists of two different types of crowd flows, including new-flow and end-flow \cite{Hoang2016}. 
Here, we adopt a total of 4-residual-unit ST-ResNet, and consider the metadata as external features like DeepST \cite{Zhang2016}. 
ST-ResNet has relatively from $14.8\%$ up to $37.1\%$ lower RMSE than these baselines, demonstrating that our proposed model has good generalization performance on other flow prediction tasks. 

\begin{table}[!htbp]
\tabcolsep 8pt
\begin{center}
\caption{Comparisons with baselines on BikeNYC. The results of ARIMA, SARIMA, VAR and 4 DeepST variants are taken from \cite{Zhang2016}. }
\label{tab:BikeNYC} 
\begin{tabular}{l|r}
\hline
Model & RMSE\\
\hline
ARIMA & 10.07\\
SARIMA	 & 10.56\\
VAR	 & 9.92\\
DeepST-C	& 8.39\\
DeepST-CP 	& 7.64\\
DeepST-CPT	& 7.56\\
DeepST-CPTM & 	7.43\\
\hline
ST-ResNet [ours, 4 residual units]	&  \textbf{6.33}\\
\hline
\end{tabular}
\end{center}
\end{table}

\section{Related Work}
\textbf{Crowd Flow Prediction.} There are some previously published works on predicting an individual's movement based on their location history \cite{Fan2015,Song2014}. 
They mainly forecast millions, even billions, of individuals' mobility traces rather than the aggregated crowd flows in a region. 
Such a task may require huge computational resources, and it is not always necessary for the application scenario of public safety. 
Some other researchers aim to predict travel speed and traffic volume on the road \cite{Abadi2015IToITS,Silva2015PotNAoS,Xu2014IToITS}. Most of them are predicting single or multiple road segments, rather than citywide ones. 
Recently, researchers have started to focus on city-scale traffic flow prediction \cite{Hoang2016,Li2015}. 
Both work are different from ours where the proposed methods naturally focus on the individual region not the city, and they do not partition the city using a grid-based method which needs a more complex method to find irregular regions first. 

\noindent\textbf{Deep Learning.} CNNs have been successfully applied to various problems, especially in the field of computer vision \cite{Krizhevsky2012}. Residual learning \cite{He2015apa} allows such networks to have a very super deep structure. Recurrent neural networks (RNNs) have been used successfully for sequence learning tasks \cite{sutskever2014sequence}. The incorporation of long short-term memory (LSTM) enables RNNs to learn long-term temporal dependency. However, both kinds of neural networks can only capture spatial or temporal dependencies. Recently, researchers combined above networks and proposed a convolutional LSTM network \cite{xingjian2015convolutional} that learns spatial and temporal dependencies simultaneously. Such a network cannot model very long-range temporal dependencies (\eg{}, period and trend), and training becomes more difficult as depth increases. 

In our previous work \cite{Zhang2016}, a general prediction model based on DNNs was proposed for spatio-temporal data. In this paper, to model a specific spatio-temporal prediction (\ie{} citywide crowd flows) effectively, we mainly propose employing the residual learning and a parametric-matrix-based fusion mechanism. A survey on data fusion methodologies can be found at \cite{Zheng2015Itobd}. 

\section{Conclusion and Future Work}
We propose a novel deep-learning-based model for forecasting the flow of crowds in each and every region of a city, based on historical trajectory data, weather and events. 
We evaluate our model on two types of crowd flows in Beijing and NYC, achieving performances which are significantly beyond 6 baseline methods, confirming that our model is better and more applicable to the crowd flow prediction. The code and datasets have been
released at: {https://www.microsoft.com/en-us/research/publication/deep-spatio-temporal-residual-networks-for-citywide-crowd-flows-prediction}.

In the future, we will consider other types of flows (\eg{}, taxi/truck/bus trajectory data, phone signals data, metro card swiping data), and use all of them to generate more types of flow predictions, and \textit{collectively} predict all of these flows with an appropriate fusion mechanism. 

{
\small
\bibliographystyle{aaai}
\bibliography{ref}
}
\end{document}